\documentclass[a4paper, 10pt, conference]{ieeeconf}      
\usepackage{FG2025}

\FGfinalcopy 

\IEEEoverridecommandlockouts                              
\usepackage{graphicx} 
\usepackage{amsmath} 
\usepackage{amssymb}  
\usepackage{algorithm}  
\usepackage{algorithmic}

\def\FGPaperID{0009} 

\title{\LARGE \bf
Regression Guided Strategy to Automated Facial Beauty Optimization through Image Synthesis
}

\author{\parbox{16cm}{\centering
    {\large Erik Nguyen$^1$ and Spencer Htin$^2$}\\
    {\normalsize
    $^1$ Department of Computer Science, University of California San Diego, USA\\
    $^2$ Fairmont Preparatory Academy, USA}}
}

\begin{document}

\ifFGfinal
\thispagestyle{empty}
\pagestyle{empty}
\else
\author{Anonymous FG2025 submission\\ Paper ID \FGPaperID \\}
\pagestyle{plain}
\fi
\maketitle

\begin{abstract}

The use of beauty filters on social media, which enhance the appearance of individuals in images, is a well-researched area, with existing methods proving to be highly effective. Traditionally, such enhancements are performed using rule-based approaches that leverage domain knowledge of facial features associated with attractiveness, applying very specific transformations to maximize these attributes. In this work, we present an alternative approach that projects facial images as points on the latent space of a pre-trained GAN, which are then optimized to produce beautiful faces. The movement of the latent points is guided by a newly developed facial beauty evaluation regression network, which learns to distinguish attractive facial features, outperforming many existing facial beauty evaluation models in this domain. By using this data-driven approach, our method can automatically capture holistic patterns in beauty directly from data rather than relying on predefined rules, enabling more dynamic and potentially broader applications of facial beauty editing. This work demonstrates a potential new direction for automated aesthetic enhancement, offering a complementary alternative to existing methods.

\end{abstract}

\section{INTRODUCTION}

Facial beauty filters have become ubiquitous in social media and photo-editing applications, allowing users to easily enhance their appearance. These filters typically rely on rule-based methods, which use predefined transformations based on established beauty ideals, such as proportionality, eye size, skin smoothness, and other specific feature adjustments. While shown to be extremely effective for enhancing beauty, these methods are constrained by their reliance on static rules. 

This raises a compelling research question: Instead of relying on domain knowledge or specific feature adjustments, what if we could create a system that automatically learns what constitutes beauty in faces, and then edits a photo based on what it has learned? 

In this work, we explore an alternative approach to enhancing facial aesthetics in images by leveraging machine learning and generative models. Guided by a facial beauty evaluation regression model, our method optimizes an objective function within the latent space of a GAN, enabling precise and effective image editing.

\subsection{Related Works}
Beauty enhancement in digital imagery has been a popular topic of research for years, leading to a variety of approaches aimed at improving facial aesthetics. One such approach is to use Generative Adversarial Networks (GAN) \cite{goodfellow2014generative}, as described by Liu \textit{et al.} \cite{10194986}. Here, a GAN was used to manipulate specific facial attributes like mouth expression, eyebrow position, and facial hair. Similarly, others such as Li \textit{et al.} \cite{10.1145/3240508.3240618} and Chen \textit{et al.} \cite{Chen_2019_CVPR} also approached this problem using a GAN, but employed a style transfer technique to apply makeup from reference images, enhancing perceived beauty through the transfer of makeup styles.

Most related to our approach is Style-based Age Manipulation (SAM) by Alaluf \textit{et al.} \cite{alaluf2021only}, which uses an age regression model to guide the training of an encoder for the generation of aged faces. In our strategy, we develop a facial beauty evaluation regression network that serves as our guiding model, steering latent point optimization to achieve enhanced beauty.


\begin{figure*}[t]
    \centering
    \hspace*{-0.7cm}
    \includegraphics[width=\linewidth]{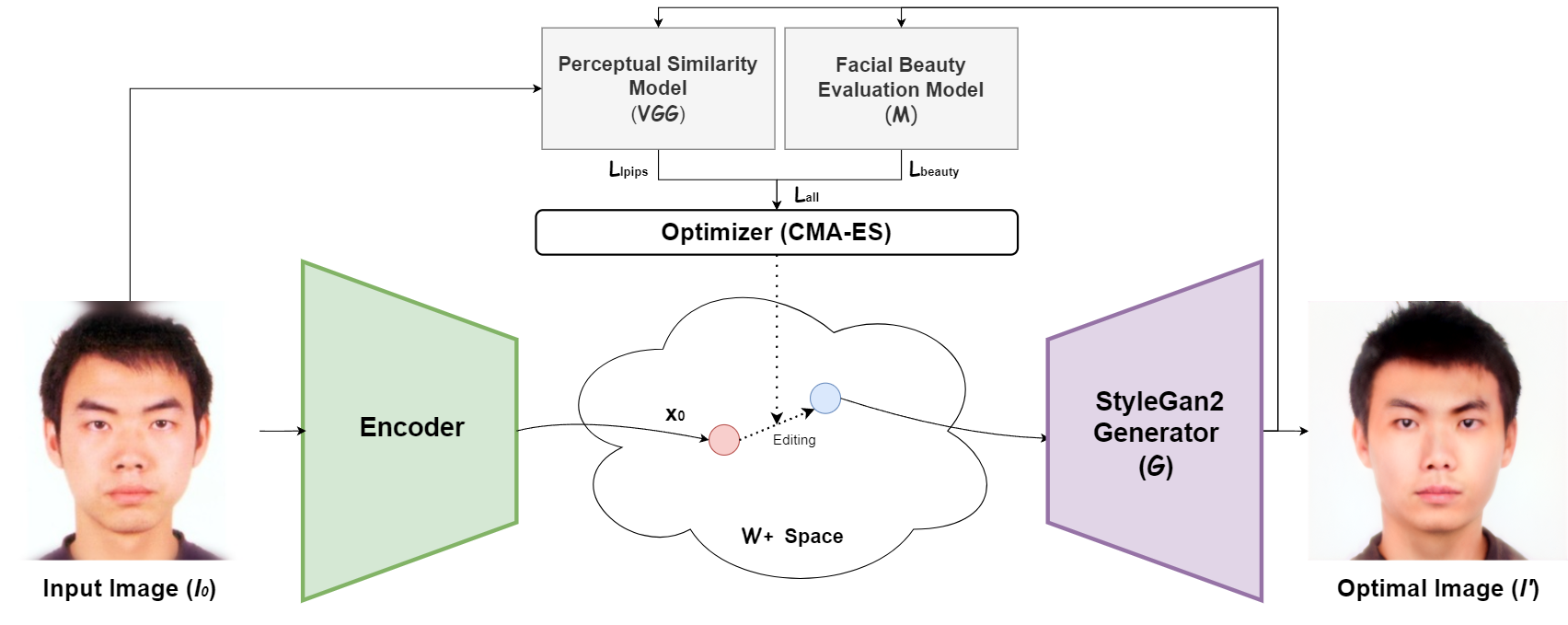} 
    \caption{Strategy for facial aesthetics optimization through image synthesis.}
    \label{fig:architecture}
\end{figure*}


\section{METHODOLOGY}

In this section, we provide an overview of our strategy for optimizing and editing facial aesthetics given input facial images. The process involves several key stages: projecting the image into the latent space of StyleGAN2 \cite{karras2020analyzingimprovingimagequality}, optimizing for beauty using a facial beauty regression model, and finally, producing the enhanced image. The flow of this process is also visualized in Fig. \ref{fig:architecture}.

\subsection{GAN Inversion}
First, the given facial image must be projected onto the latent space of a pre-trained StyleGAN2 generator. This step is commonly known as GAN inversion, and it involves mapping the input image $I_{0}$ to a point $\in\mathbb{R}^{18\times512}$ in the $\mathcal{W+}$ latent space that approximates the original image as closely as possible when reconstructed. 

In our strategy, we utilize a hybrid approach to GAN inversion, where we first trained an encoder network, based on the ResNet-101 architecture \cite{he2016deep}, to map real facial images into the $\mathcal{W+}$ latent space of the pre-trained StyleGAN2 generator, which we denote as $\mathcal{G}$. The generator, pre-trained on the FFHQ dataset \cite{Karras_2019_CVPR}, is capable of synthesizing high-quality facial images from latent points.

\begin{equation}
proj := \text{Encoder}(I_{0})
\end{equation}

Then, the initial latent point obtained from the encoder is refined for several iterations using gradient-based optimization. We apply the Adam optimizer \cite{kingma2017adammethodstochasticoptimization} to further adjust $proj$ to minimize the perpetual loss $\mathcal{L}_{lpips}$, which ensures that the generated image is as similar as possible to the original input image in terms of perceptual features. The details of the loss function will be discussed in an upcoming section.

\begin{equation}
x_{0} := \arg\min_{proj} \mathcal{L}_{lpips}(\mathcal{G}(proj))
\end{equation}

\subsection{Beauty Optimization}
With the refined latent point $x_{0}$ obtained from GAN Inversion, we proceed to the main optimization stage using the Covariance Matrix Adaptation Evolution Strategy (CMA-ES) \cite{hansen2016cma}. This was chosen over gradient-based optimizers like the previously used Adam optimizer due to the noisy outputs of our facial beauty estimation model, which was found to hinder convergence in gradient-based methods.

Here, the initial step size $\sigma_{0}$ is set to $0.06$, which is a hyperparameter in CMA-ES that controls how far the samples can deviate from the initial latent point $x_{0}$ during optimization, allowing the algorithm to explore the latent space effectively while also staying close enough to the original position.

The goal of this optimization is to find the optimal latent point $x_{optimal}$ that produces a maximally beautiful face while also preserving as much of the original facial features as possible. This optimization step is run for a few hundred iterations, where the number of iterations controls how much the output image is refined.

\begin{equation}
x_{optimal} := \arg\min_{x} \mathcal{L}_{all}(\mathcal{G}(x))
\end{equation}
where $\mathcal{L}_{all}$ is the combined loss function that we define in the next section.

\subsection{Loss Functions}
To guide the optimization and ensure that the generated image is both beautiful and perceptually similar to the input image, we define the following loss functions:

\subsubsection{Perceptual Loss ($\mathcal{L}_{lpips}$)}
This loss function measures the perceptual similarity between the input image $I_{0}$ and the generated image $I$. It is computed using a pre-trained VGG-16 model, which is known for its effectiveness in capturing perceptual features relevant to human vision \cite{zhang2018unreasonable}. The VGG model extracts high-level feature representations from images, which are then compared using the following:

\begin{equation}
\mathcal{L}_{lpips}(I) = \| \text{VGG}(I_{0}) - \text{VGG}(I) \|^2
\end{equation}

\subsubsection{Beauty Loss ($\mathcal{L}_{beauty}$)}
This loss function quantifies how well the generated image meets the beauty criteria, which was automatically learned from our facial beauty evaluation regression model $\mathcal{M}$. The model was trained to score facial images within a fixed range of 1 to 5, where a lower number represents a lower amount of beauty. Below, the constant $c$ represents the highest possible beauty score (5) that the model can output. 

\begin{equation}
\mathcal{L}_{beauty}(I) = (c - \mathcal{M}(I))^2
\end{equation}

\subsubsection{Combined Loss ($\mathcal{L}_{all}$)}
The overall objective is to balance facial aesthetic transformations and perceptual similarity of the generated image. The combined loss function is defined as:

\begin{equation}
\mathcal{L}_{all}(I) = \beta_{1} \max(\mathcal{L}_{lpips}(I), \theta) + \beta_{2} \mathcal{L}_{beauty}(I)
\end{equation}
where $\beta_{1}$ and $\beta_{2}$ are weights that control the trade-off between maintaining the original facial features and enhancing beauty. A higher $\beta_{1}$ prioritizes preserving the original facial features, while a higher $\beta_{2}$ controls how much the facial features can be modified to enhance beauty. Threshold $\theta$ works in a similar way by limiting the similarity to the original image.

\subsection{Final Image Generation}
After obtaining the latent point $x_{optimal}$ using CMA-ES, we use the StyleGan2 generator to produce the final image:

\begin{equation}
I' := \mathcal{G}(x_{optimal})
\end{equation}
This image $I'$ is the result of our process and represents the maximally beautiful version of the original facial image according to the defined criteria learned from the facial beauty evaluation regression model.


\begin{figure}[t]
    \centering
    \includegraphics[scale=0.3]{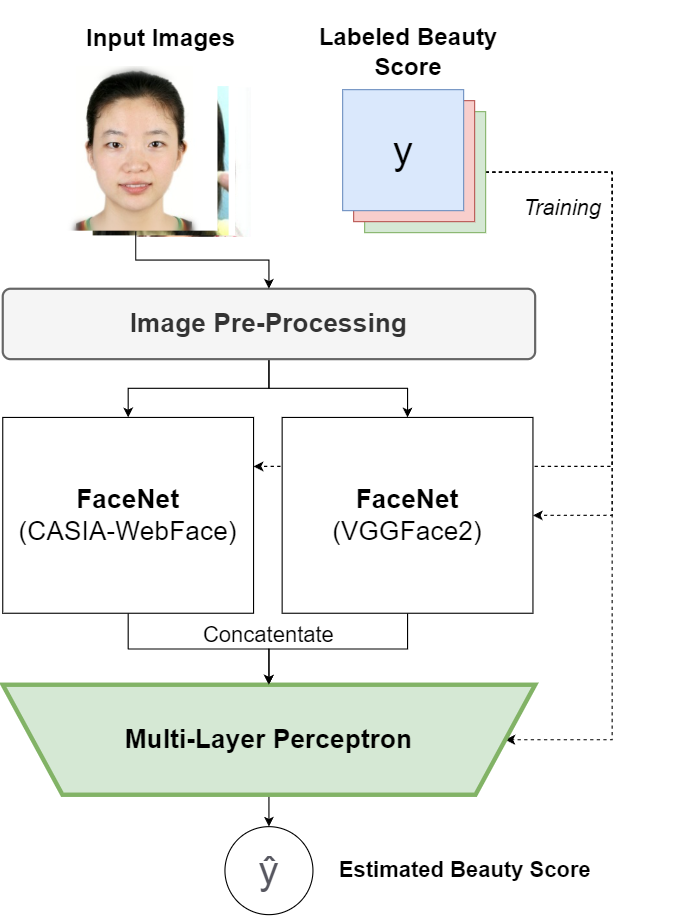}
    \caption{FaceNet Ensemble model for facial beauty evaluation.}
    \label{fig:fbp}
\end{figure}


\section{Facial Beauty Evaluation Model}
Guidance for the optimization process in generating maximally beautiful facial images is the most crucial aspect of our strategy. To effectively steer this process, we developed a facial beauty evaluation model that predicts beauty scores for facial images, providing feedback during optimization, and ensuring that the generated images align with the desired increase in perceived beauty.

We frame the facial beauty evaluation task as a supervised learning regression problem, aiming to train a model that predicts a continuous attractiveness score for each facial image. For this, we utilize the SCUT-FBP5500 dataset \cite{liang2018scutfbp5500}, which is comprised of 5500 images of Caucasian and Asian men and women. Each image is labeled with an average beauty score, derived from assessments by 60 volunteers, with scores ranging from 1 (low beauty) to 5 (high beauty).

\subsection{Image Pre-Processing \& Model Architecture}
Our model pipeline starts with aligning and transforming images using the same procedures for image preparation in the FFHQ dataset. 

The aligned images are processed through two FaceNet models \cite{schroff2015facenet}, both using the InceptionResnetV1 architecture. One model was pre-trained on CASIA-WebFace dataset \cite{yi2014learning}, and the other on VGGFace2 \cite{8373813}, with each producing a high-dimensional embedding $\in\mathbb{R}^{512}$. These embeddings are concatenated into a single feature vector, which is then passed through a multi-layer perceptron (MLP) with Mish activation \cite{misra2019mish} to produce the final beauty score. An illustration of the model's architecture is shown in Fig. \ref{fig:fbp}.

\subsection{Training}
We fine-tuned the FaceNet models by freezing the top layers and unfreezing the last few. The MLP was trained in its entirety. This approach allowed the models to adapt their high-level feature representations specifically for the task of beauty evaluation while preserving their foundational, lower-level understanding of facial features.

The combined model was trained for 100 epochs, utilizing the Ranger optimizer \cite{Ranger} along with a cosine annealing learning rate scheduler with linear warmup to enhance convergence and stability when updating the model parameters.

\subsection{Evaluation}
To evaluate how well the model performed, we conducted 5-fold cross-validation. The model's performance was evaluated using the standard regression metrics mean absolute error (MAE), root mean squared error (RMSE), and Pearson correlation coefficient (PC). Table \ref{tab:performance1} compares the model's performance metrics with other studies, showing that our model matches or surpasses most of the current state-of-the-art models in this domain, proving its effectiveness for the task of facial beauty evaluation.


\begin{table}[t]
    \centering
    \caption{Model Performance Metrics on The SCUT-FBP5500 Dataset with 5-fold Cross Validation}
    \begin{tabular}{clll}
        \hline
        Model & $\downarrow$ MAE & $\downarrow$ RMSE & $\uparrow$ PC \\
        \hline
        Gabor PCA + GR \cite{liang2018scutfbp5500} & $0.355$ & $0.459$ & $0.747$ \\
        GNN-SVR \cite{nguyen2024racialbeautymodeling} & $0.273$ & $0.361$ & $0.851$ \\
        AlexNet \cite{liang2018scutfbp5500} & $0.265$ & $0.348$ & $0.863$ \\
        ResNet-18 \cite{liang2018scutfbp5500} & $0.241$ & $0.316$ & $0.890$ \\
        ResNeXt-50 \cite{liang2018scutfbp5500} & $0.2293$ & $0.316$ & $0.899$ \\
        CNN-ER \cite{BOUGOURZI2022108246} & $0.200$ & $0.265$ & $0.925$ \\
        NMFA \cite{10065499} & $\textbf{0.178}$ & $0.261$ & $0.923$ \\
        FaceNet Ensemble & $0.188$ & $\textbf{0.252}$ & $\textbf{0.931}$ \\
        \hline
    \end{tabular}
    \label{tab:performance1}
\end{table}


\begin{figure*}[t]
    \centering
    \includegraphics[width=\textwidth]{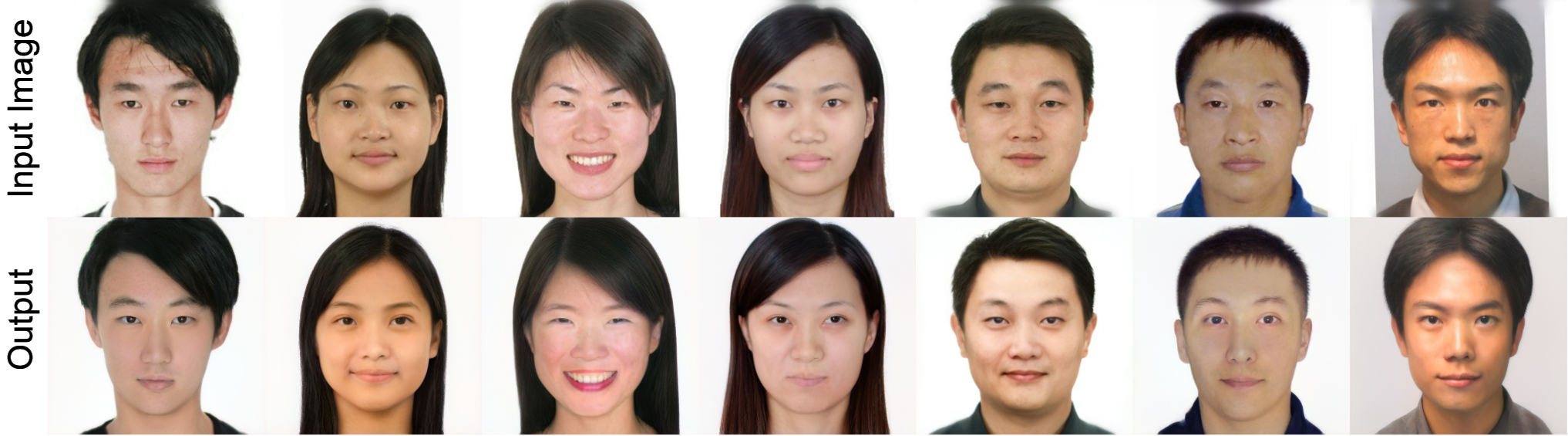}
    \caption{Results generated using the beauty optimization strategy.}
    \label{fig:examples}
\end{figure*}


\section{Results}
Comparison between the original and generated facial images shows that our method effectively enhanced facial aesthetics. As shown in Fig. \ref{fig:examples}, the improvements are visually noticeable. This indicates that our strategy effectively captured complex beauty patterns and applied them successfully, without relying on predefined rules or targeted feature adjustments.

\subsection{Observed Attribute Changes}

Upon closer examination, changes to several specific facial attributes were observed across the output images:
\begin{itemize}
    \item{\textit{Skin Texture and Clarity}: A significant improvement in skin clarity was observed in almost all of the output images. These modifications seemed to reduce wrinkles or other imperfections, giving the skin a healthier look.}
    \item{\textit{Eyebrows}: Some output images featured alterations in the tilt and position of the eyebrows.}
    \item{\textit{Chin Definition}: In some of the images, the chin and jawline were made slightly more defined.}
    \item{\textit{Hair Refinement}: The hair in many of the photos appeared neater and more uniformly styled, with adjustments that made it straighter or smoother.}
    \item{\textit{Lips}: Some images showed modifications to the lips, including adjustments to lip thickness, shape, and color.}
\end{itemize}

\subsection{Discussion}



A notable result of our strategy is its ability to make multiple changes at once, affecting various facial attributes simultaneously. This characteristic agrees with the holistic concept of beauty, where attractiveness is perceived not as a sum of isolated features, but as the harmonious combination of multiple facial attributes. Our strategy's enhancements reflect this holistic view by simultaneously optimizing for those learned attributes from the guiding model.


\begin{figure}[ht]
    \centering
    \includegraphics[width=\linewidth]{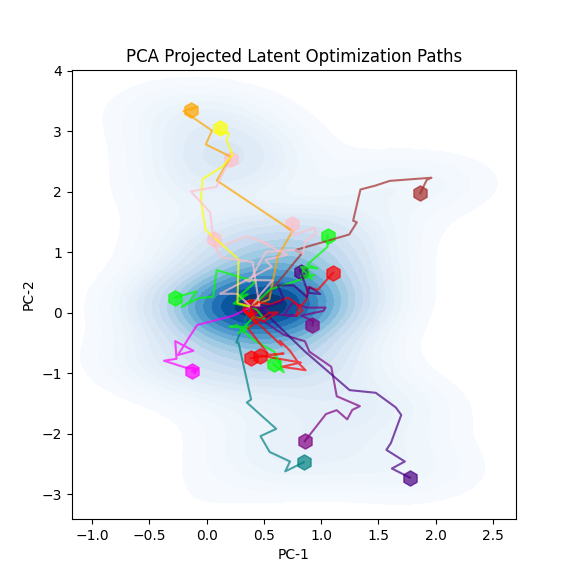}
    \caption{Optimization paths in latent space projected via PCA.}
    \label{fig:paths}
\end{figure}


As a result, the optimization paths in latent space exhibit a high degree of entanglement, as illustrated in Fig. \ref{fig:paths}, showing complex and non-uniform trajectories as it optimizes across these numerous attributes.

While this entanglement poses a significant challenge in identifying clear patterns or directions, it also opens up an intriguing avenue for future research. Future work can focus on disentangling these paths, which would provide clearer insights into the optimization process, and make the learned facial patterns from the guiding model interpretable.

\subsection{Limitations}
Due to a lack of more diverse datasets, we used a public dataset predominantly labeled by East Asian individuals to train the beauty evaluation guiding model, which may have led the model to reflect East Asian beauty standards, such as fairer skin and neater hair. This cultural bias likely limits the generalizability of our results, which may not represent diverse beauty ideals across different populations.

Another limitation is that this method relies on the pre-trained StyleGAN for face editing. Certain face archetypes or features that are underrepresented in the training data of StyleGAN may pose challenges for GAN inversion, leading to less accurate reconstructions and edits. This can result in suboptimal outcomes for individuals whose facial characteristics deviate significantly from the distribution of faces.





\section{CONCLUSION}

In this study, we introduce a strategy for facial beauty enhancement that utilizes regression-guided optimization. Unlike traditional rule-based approaches, which rely on domain knowledge or specific feature adjustments, our method leverages a pre-trained GAN and a newly developed facial beauty evaluation regression model to learn and apply complex patterns associated with facial beauty. Here, the regression model learns to identify and quantify beauty, which guides the optimization process within the latent space of the GAN. This approach avoids the constraints of static rules and enables more dynamic and holistic adjustments. 

Our results, despite the method being extremely simple, confirm that this strategy effectively enhances facial aesthetics by simultaneously optimizing multiple facial attributes in a holistic manner. This method offers a complementary alternative to existing enhancement techniques, paving the way for potentially more flexible and personalized applications in beauty editing.

{\small
\bibliographystyle{ieee}
\bibliography{egbib}
}

\end{document}